\documentclass[conference]{IEEEtran}
\IEEEoverridecommandlockouts
\usepackage{cite}
\usepackage{amsmath,amssymb,amsfonts}
\usepackage{algorithmic}
\usepackage{graphicx}
\usepackage{textcomp}
\usepackage{xcolor}
\usepackage{booktabs}
\usepackage[numbers]{natbib}

\usepackage{float}
\usepackage{url}
\usepackage{multirow}

\def\BibTeX{{\rm B\kern-.05em{\sc i\kern-.025em b}\kern-.08em
    T\kern-.1667em\lower.7ex\hbox{E}\kern-.125emX}}
\begin{document}

\title{Enabling On-Device Medical AI Assistants via Input-Driven Saliency Adaptation
}
\title{Enabling On-Device Medical AI Assistants via Input-Driven Saliency Adaptation}

\author{
    Uttej Kallakurik, Edward Humes, Rithvik Jonna, Xiaomin Lin, Tinoosh Mohsenin\\
    Johns Hopkins Whiting School of Engineering, Baltimore, Maryland, United States\\
    \texttt{\{ukallak1, ehumes2, djonna1, xlin52, tinoosh\}@jhu.edu}
    \thanks{Accepted for publication at IEEE BioCAS 2025.}
}

\maketitle

\begin{abstract}
Large Language Models (LLMs) have significant impact on the healthcare scenarios but remain prohibitively large for deployment in real-time, resource-constrained environments such as edge devices. In this work, we introduce a novel medical assistant system, optimized through our general-purpose compression framework, which tailors Large Language Models (LLMs) for deployment in specialized domains.
By measuring neuron saliency on domain-specific data, our method can aggressively prune irrelevant neurons, reducing model size while preserving performance.
Following pruning, we apply post-training quantization to further reduce the memory footprint, and evaluate the compressed model across medical benchmarks including MedMCQA, MedQA, and PubMedQA. We also deploy the 50\% compressed Gemma and the 67\% compressed LLaMA3 models on Jetson Orin Nano (18.7W peak) and Raspberry Pi 5 (6.3W peak), achieving real-time, energy-efficient inference under hardware constraints.
\end{abstract}

\section{Introduction}

Large Language Models (LLMs) have rapidly advanced the capabilities of natural language understanding\cite{li2025benchmark}, generation\cite{gatt2018survey}, and reasoning\cite{yu2024natural}, establishing themselves as critical tools across diverse applications—such as general conversation\cite{cai2025natural}, conversation guided simulation\cite{palnitkar2023chatsim}, content understanding\cite{mao2025multi} and domain-specific question answering\cite{ jin2021disease, jin2019pubmedqa}. 
These billion-parameter models are highly effective but too resource intensive for real-time use on edge devices in clinical settings or mobile health applications.
In medical applications, there is growing interest in using LLMs as virtual assistants for doctors and patients alike. These assistants must interpret domain-specific questions, generate accurate and grounded responses, and run autonomously in environments where privacy, latency, and power consumption are critical. However, deploying full-scale LLMs on edge devices remains extremely difficult without aggressive compression techniques that preserve task-relevant performance while meeting tight hardware constraints.

\begin{figure}[ht]
    \centerline{\includegraphics[width=0.5\textwidth]{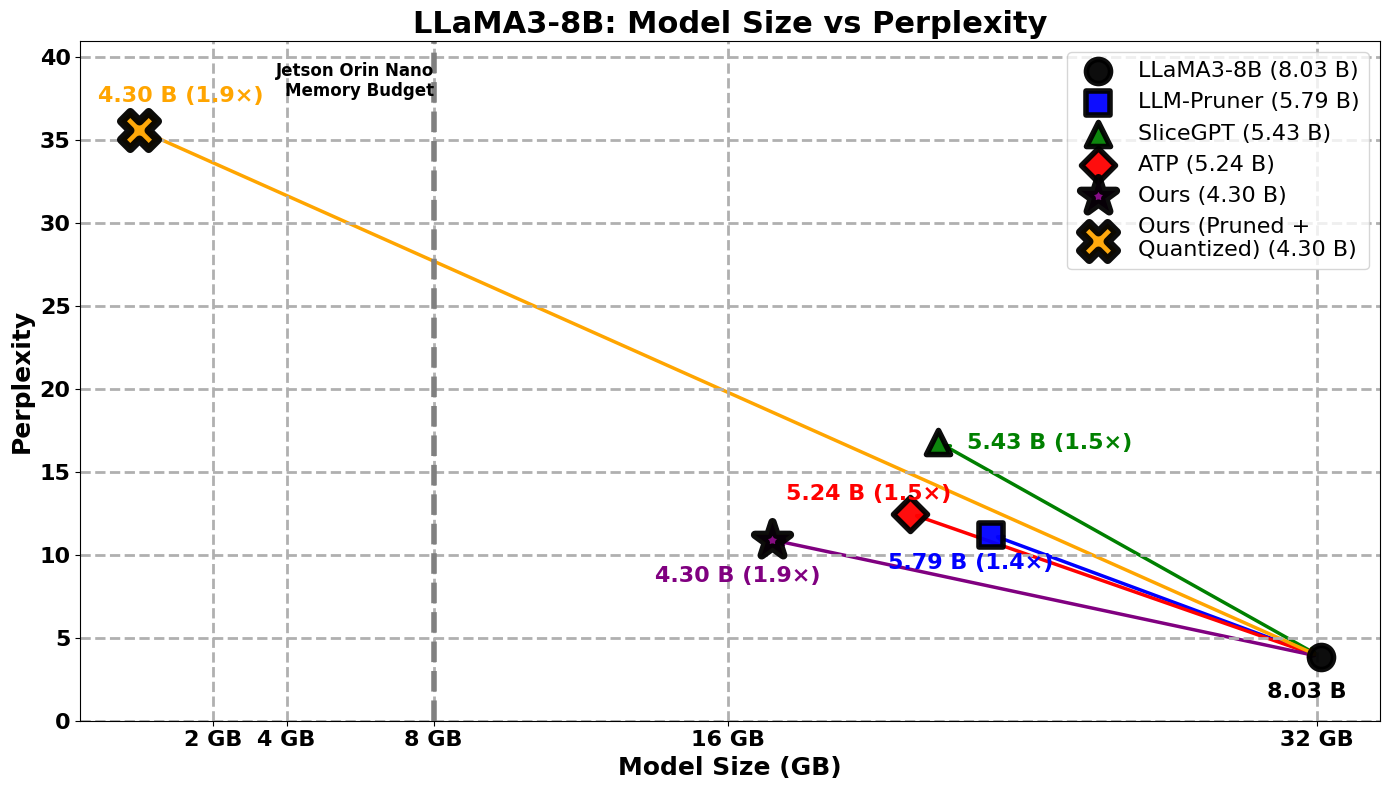}}
    \vspace{-2ex}
    \caption{\footnotesize
    Perplexity vs. model size (in GB, 32-bit equivalent) for LLaMA3-8B and its pruned variants. Our method (purple star) achieves the lowest perplexity (10.9) among all compressed models while reducing parameter count to 4.30B. Compared to LLM-Pruner\cite{ma2023llm}, SliceGPT\cite{ashkboos2024slicegpt}, and ATP\cite{lu2024allinonetuningstructuralpruning}, our approach achieves up to 1.9$\times$ lower perplexity at a smaller model size. This demonstrates the effectiveness of task-specific pruning in preserving performance under aggressive compression.
    }
    \vspace{-4ex}
    \label{fig:llama3_perplexity}
\end{figure}
\vspace{1ex}

To address the challenge of deploying LLMs in clinical settings, we present a task-specific compression framework that adapts to domain-relevant data. Crucially, neuron saliency in transformer models varies with input distribution; by leveraging datasets like Medical Meadow~\cite{han2023medalpaca}, we identify neurons most critical for medical language tasks. This input-driven pruning exposes and retains clinically relevant neurons while safely removing inactive ones, enabling significant model compression with minimal loss in task-specific performance.

Fig.~\ref{fig:llama3_perplexity} illustrates the trade-off between model size and perplexity for the LLaMA3-8B (B here stands for Billions of parameters) model \cite{grattafiori2024llama} and its pruned variants under various state-of-the-art compression techniques. Our proposed method (purple star) achieves a substantial reduction in model size—from 8.03B parameters down to 4.30B—while maintaining a low perplexity score of 10.9. All perplexity measurements were computed using the “Harrison Score,” derived from the \textit{Principles of Internal Medicine}~\cite{harrison1954principles}, to ensure consistency with existing pruning benchmarks. 
Compared to LLM-Pruner\cite{ma2023llm}, SliceGPT~\cite{ashkboos2024slicegpt}, and ATP~\cite{lu2024allinonetuningstructuralpruning}, our approach uses 1.2–1.6× fewer parameters and yields 1.4–1.6× lower perplexity, validating the advantage of input-driven, task-specific pruning for efficient clinical language modeling.


Our key contributions for this work are summarized as follows:
\begin{itemize}
    \item We propose a medical task-specific LLM pruning framework based on input-driven saliency adaptation, which uses domain-relevant datasets to identify neurons critical to the target application.
    \item We apply a dual-stage saliency measure, combining the L2-norm and the Jacobian norm sensitivity, to a medical language modeling task to guide pruning and integrate post-training quantization toolchains to further compress the pruned models.
    \item We benchmark the resulting compressed LLM on different medical benchmarks and deploy it on two edge hardware devices— NVIDIA's Jetson Orin Nano and a Rasberry-Pi 5 with 8GB 16GB of RAM respectively.
\end{itemize}

\section{Related Work}

\subsection{LLMs for Medical Applications}

Large Language Models (LLMs) have shown significant promise in the medical domain, particularly in tasks like medical question answering. Notably, models such as Med-PaLM~\cite{singhal2023large} and its successor Med-PaLM 2~\cite{singhal2023toward} have achieved expert-level performance on benchmarks like MedQA and PubMedQA~\cite{singhal2023large, singhal2023toward}. Similarly, BioGPT~\cite{luo2022biogpt} has demonstrated strong capabilities in biomedical text generation and mining. Despite these advancements, these models' substantial size and computational demands pose challenges for deployment in resource-constrained, privacy-sensitive environments, such as edge devices in clinical settings.

\subsection{Compression and Pruning of LLMs}

To address the deployment challenges of large LLMs, various model compression techniques have been explored. Structured pruning methods, which remove entire components like attention heads or feed-forward network (FFN) layers, have been effective in reducing model size while maintaining performance~ \cite{lu2024allinonetuningstructuralpruning, ashkboos2024slicegpt, ma2023llm}. Recent approaches, such as EfficientLLM~\cite{guo2025efficientllm}, integrate pruning into the pretraining phase, resulting in models optimized for edge deployment. However, many existing pruning strategies are task-agnostic and do not account for domain-specific input distributions, potentially limiting their effectiveness in specialized applications like medical NLP.

\subsection{Quantization and Edge Deployment}
Quantization techniques, particularly post-training quantization (PTQ)\cite{yao2023comprehensive}, have been employed to further reduce the memory footprint and computational requirements of LLMs. Tools like NVIDIA's TensorRT and the ONNX Runtime facilitate the deployment of quantized models on edge devices~\cite{nvidia2024ptq, apxml2025onnx}. While these methods enable efficient inference, there remains a need for comprehensive evaluations of quantized LLMs in real-world, domain-specific scenarios.

\section{Proposed Approach}

\begin{figure}[ht]
    \centerline{\includegraphics[width=0.5\textwidth]{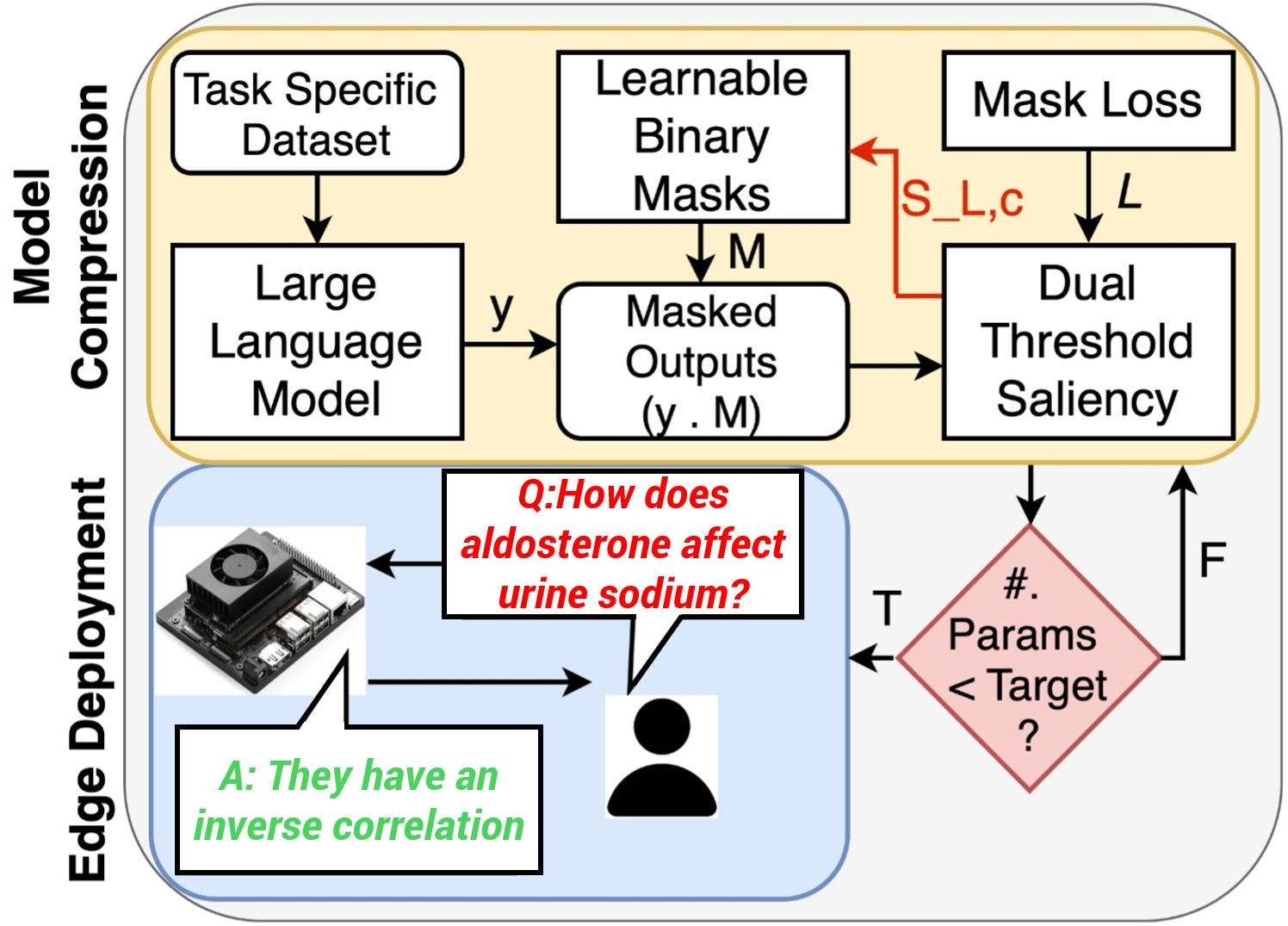}}
    \vspace{-2ex}
    \caption{\footnotesize High-level architecture, involves computing saliency measures for each neuron, which are then used to generate soft masks. We further derive binary masks from the soft masks which are then applied to the outputs of the model. A '0' on the binary mask indicates the neuron is turned-off while a '1' indicates the neuron is turned-on. The masks are updated iteratively, allowing the model to retain high accuracy while reducing complexity. Red arrows show the gradients calculated to update the Saliency thresholds while the black arrows show the computations to generate the binary masks.}
    \vspace{-4ex}
    \label{fig:biocas_2}
\end{figure}
\vspace{1ex}

\subsection{Problem Motivation and Postulation}

Large Language Models (LLMs) contain massive number of parameters that enables generalization across a wide range of tasks. However, for any specific downstream application, such as medical question answering—only a subset of the full model capacity is typically exercised during inference. This motivates our central postulation:

\begin{quote}
\textit{For application-specific tasks, we can aggressively prune large language models with minimal degradation in task-specific performance, as only a smaller subset of neurons exhibit high saliency for the domain-relevant input distribution.}
\end{quote}
Figure~\ref{fig:biocas_2} provides an overview of our architecture, where neuron saliency is used to generate binary masks that guide pruning while maintaining task-specific accuracy.

To formalize this postulation, consider a pre-trained transformer-based LLM represented as a directed acyclic graph (DAG) \( G = (V, E) \), where:
\begin{itemize}
    \item \( V \) is the set of neurons (nodes), and
    \item \( E \) is the set of weighted connections (edges) between them.
\end{itemize}

Given an input dataset \( \mathcal{D}_{\text{task}} \) corresponding to a specific application (e.g., medical QA), the saliency of each neuron \( v_i \in V \) is a function of the input distribution:
\[
\text{Saliency}(v_i) = \mathbb{E}_{x \sim \mathcal{D}_{\text{task}}} \left[ \| \text{activation}_{v_i}(x) \|_2 \right] + \| \mathbf{J}_{v_i}(x) \|_2
\]
where \( \mathbf{J}_{v_i}(x) \) denotes the Jacobian of the neuron’s output with respect to its input. Since the dataset \( \mathcal{D}_{\text{task}} \) is focused on a narrow domain, only a subset \( V' \subset V \) of neurons receive consistently high saliency values. The rest remain under-activated and can be pruned.

\subsection{Pruning as Subgraph Selection}

Our goal is to identify a subgraph \( G' = (V', E') \subset G \) that retains only the task-relevant neurons and weights. We define the task-specific pruning objective as:
\[
\min_{G' \subset G} |E'| \quad \text{subject to} \quad P(G') \geq \eta \cdot P(G)
\]
where:
\begin{itemize}
    \item \( P(G) \) is the original model's performance on the task.
    \item \( P(G') \) is the pruned model's performance.
    \item \( \eta \in [0,1] \) defines the acceptable performance retention ratio.
\end{itemize}

We perform this pruning exclusively within the feed-forward network (FFN) layers, which are known to contain the majority of the parameters and dominate inference latency.

\subsection{Input-Driven Saliency Estimation}

To estimate neuron saliency, we employ a dual-stage metric that evaluates:
\begin{enumerate}
    \item \textbf{Magnitude Saliency}: Measured using the L2 norm of the neuron output.
    \[
    S^{\text{mag}}_{i} = \| \text{activation}_{v_i}(x) \|_2
    \]
    \item \textbf{Gradient Saliency}: Measured using the Jacobian of the activation function.
    \[
    S^{\text{jac}}_{i} = \left\| \frac{\partial f(v_i)}{\partial z_i} \right\|_2
    \]
\end{enumerate}

These saliency scores are computed using a small number of unlabeled samples from the application-specific dataset \( \mathcal{D}_{\text{task}} \), such as MedQA\cite{jin2021disease}. Neurons with both low magnitude and low Jacobian saliency are deemed redundant and are removed using a differentiable binary masking function.

\subsection{Differentiable Masking and Loss Formulation}

Let \( M_i \in \{0,1\} \) be the binary mask associated with neuron \( v_i \). The effective output of the pruned network becomes:
\[
\hat{y} = f(x; M \odot W)
\]
where \( W \) is the original weight matrix, and \( \odot \) denotes element-wise masking.

To guide the pruning process, we define a resource-constrained loss:
\begin{equation}
\mathcal{L} = \alpha \cdot \text{TaskLoss} + (1 - \alpha) \cdot \text{ResourceLoss}
\end{equation}
where
\begin{equation}
\text{TaskLoss} = \left( \frac{P(G')}{P(G)} \right)^2,
\end{equation}
\begin{equation}
\text{ResourceLoss} = \left( \frac{|E'|}{|E|} - \text{TargetFraction} \right)^2.
\end{equation}

This loss is minimized by updating both the binary masks and their corresponding pruning thresholds via backpropagation.

\subsection{Quantization and Deployment}

After pruning, we apply post-training quantization using standard toolchains (e.g. llama.cpp~\cite{Gerganov2025ggml}), targeting 4-bit precision for weights. The resulting model is then compiled and deployed on two embedded platforms:
\begin{itemize}
    \item \textbf{Jetson Orin Nano}: A GPU-based edge platform.
    \item \textbf{Raspberry Pi 5}: A low-cost CPU-based edge device.
\end{itemize}
\vspace{1ex}

The compressed model is evaluated for latency, memory usage, and energy efficiency on both platforms to validate its suitability for real-time, privacy-preserving medical assistant applications.

\section{Results}

\subsection{Experimental Setup}

In this section, we evaluate our pruning+fine‐tuning pipeline on two large language models—Gemma\,1\,7B~\cite{team2024gemma} and Llama\,3\,8B~\cite{grattafiori2024llama}. Our experimental pipeline consists of three main phases: \textit{(i)} pruning through input-driven saliency estimation on a medical domain corpus, \textit{(ii)} task-specific fine-tuning, and \textit{(iii)} evaluation on benchmark medical datasets.
We apply a dual-stage saliency mechanism to identify task-relevant neurons for pruning, combining activation magnitude and Jacobian sensitivity. Using the Medical Meadow dataset~\cite{han2023medalpaca} ensures saliency reflects clinical relevance. Neurons with low scores are gradually masked and pruned, yielding a compact, domain-specialized LLM.
\vspace{-2ex}
\begin{figure}[!ht]
    \centerline{\includegraphics[width=0.44\textwidth]{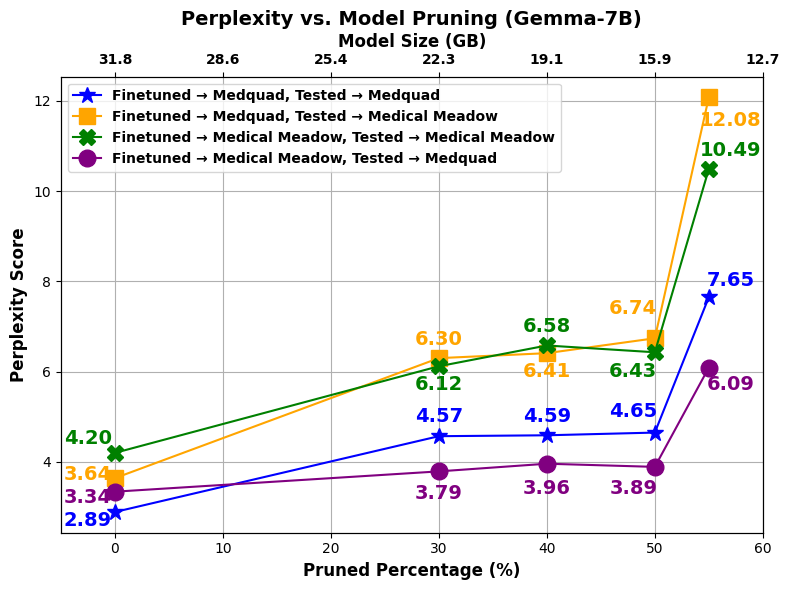}}
    \vspace{-2ex}
    \caption{\footnotesize Perplexity vs. pruning percentage for Gemma\,7B. Models are fine-tuned on either MedQuAD or Medical Meadow, then evaluated on both datasets. Perplexity remains relatively stable up to 50\% pruning, after which degradation becomes significant.}
    \vspace{-1ex}
    \label{fig:gemma_pruning}
\end{figure}
\vspace{-2ex}
\begin{figure}[!ht]
    \centerline{\includegraphics[width=0.44\textwidth]{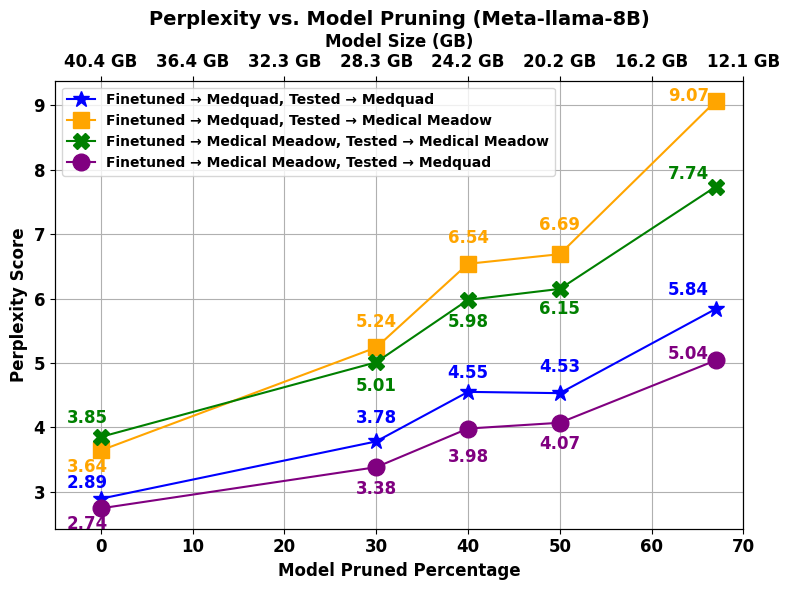}}
    \vspace{-2ex}
    \caption{\footnotesize Perplexity vs. pruning percentage for LLaMA3\,8B. Both in-domain and cross-domain performance remain robust up to 67\% pruning. Beyond this point, perplexity escalates rapidly, indicating performance collapse.}
    \vspace{-2ex}
   \label{fig:llama_pruning}
\end{figure}

After pruning, we adapt the compressed model to downstream tasks using the Low-Rank Adaptation (LoRA) technique. LoRA inserts lightweight, rank-constrained updates into the transformer’s linear layers, enabling efficient training without modifying the base weights. We produce two fine-tuned variants:
\begin{itemize}
    \item Medical Meadow Model: Fine-tuned on the Medical Meadow dataset \cite{han2023medalpaca} to retain general medical reasoning and instruction-following capability.
    \item MedQuAD Model: Fine-tuned on the MedQuAD dataset~\cite{BenAbacha-BMC-2019}, which comprises high-quality, fact-based medical QA pairs from authoritative health sources such as MedlinePlus and NIH.
\end{itemize}
\vspace{-3mm}



\subsection{Perplexity Trends vs. Pruning Rate}

To evaluate the effect of pruning on language model performance, we measure perplexity scores for Gemma\,7B and LLaMA3\,8B—each fine-tuned on the MedQuAD and Medical Meadow datasets. We report both in-domain and cross-domain perplexity across increasing pruning percentages.

Figure~\ref{fig:gemma_pruning} shows perplexity trends for the Gemma\,7B model under pruning. When fine-tuned on MedQuAD, in-domain perplexity rises moderately from 2.89 to 4.65 at 50\% pruning, while cross-domain (Medical Meadow) increases from 3.64 to 6.74. Medical Meadow–tuned models follow a similar pattern: in-domain perplexity rises from 4.20 to 6.43, and cross-domain (MedQuAD) remains low (3.34–3.89) up to 50\%, beyond which degradation accelerates.

Figure~\ref{fig:llama_pruning} shows results for the LLaMA3\,8B model. Perplexity remains stable up to 67\% pruning. For instance, the MedQuAD-tuned model increases modestly from 2.89 to 5.84 in-domain and reaches 9.07 on Medical Meadow. Similarly, the Medical Meadow–tuned variant stays below 5.1 in-domain and 5.04 cross-domain. Beyond 67\%, perplexity rises sharply, indicating degraded performance.

These trends confirm that task-specific saliency enables compression with minimal in-domain loss, though cross-domain generalization declines earlier. We therefore select the 50\% pruned Gemma\,7B and 67\% pruned LLaMA3\,8B models as deployment candidates, balancing size and performance before collapse.
\begin{table}[t]
  \centering
  \scriptsize
  \setlength{\tabcolsep}{3pt} 
  \begin{tabular}{@{}llc|ccc@{}}
    \toprule
    \textbf{Model} & \textbf{Config} & \textbf{Params (B)} 
                  & \textbf{MedMCQA} & \textbf{MedQA} & \textbf{PubMedQA} \\
    \midrule
    Gemma\,7B     & Baseline (FT)    & 9.32   & 42.98\% & 38.74\% & 72.14\% \\
    Gemma\,7B     & 50\% Pruned (FT) & 6.59    & 30.82\% & 22.20\% & 61.70\% \\
    LLaMA\,3\,8B  & Baseline (FT)    & 8.03 & 30.98\% & 20.55\% & 70.30\% \\
    LLaMA\,3\,8B  & 67\% Pruned (FT) & 5.58  & 29.45\% & 20.59\% & 43.15\% \\
    \bottomrule
  \end{tabular}
  \vspace{1mm}
  \caption{Accuracy (\%) on medical QA datasets for baseline and compressed models.}
  \vspace{-10mm}
  \label{tab:combined_results}
\end{table}

\subsection{Benchmarking Results on Medical Tasks}
To evaluate the effectiveness of our task-specific pruning and quantization pipeline, we benchmark the compressed LLMs on three medical-domain-relevant metrics using the LightEval\cite{lighteval} 
benchmarking framework. These datasets span multiple-question formats and difficulty levels, offering a comprehensive view of task-specific retention after compression. We test the 2 candidate varaints of the pruned and finetuned Gemma 7B and LLaMA 3 8B models and report results.

As shown in Table~\ref{tab:combined_results}, the compressed models maintain high task accuracy across all benchmarks, despite achieving significant reductions in model size (up to 5$\times$ smaller). For instance, the 50\% pruned and quantized Gemma 7B model achieves 30.82\% on MedMCQA and 61.7\% on PubMedQA while reducing the model size from 28\,GB to 5\,GB. Similarly, LLaMA 3 8B—when pruned by 67\% and quantized—retains comparable accuracy on MedMCQA and MedQA while achieving a drastic compression.

\section{Hardware Results}

Once pruned and fine-tuned, we selected two representative models for deployment under 4-bit quantization: the 50\% pruned Gemma\,7B and the 67\% pruned LLaMA3\,8B. These models were compact enough to fit on our target hardware after quantization. We used llama.cpp~\cite{Gerganov2025ggml} for deployment, which provides optimized CUDA and ARM backends, as well as quantization support. Specifically, we applied the Q4\_0 4-bit integer format.

\begin{figure}[!ht]
    \centering
    \vspace{-3mm}
    \begin{minipage}[t]{0.48\linewidth}
        \centering
        \includegraphics[width=\linewidth]{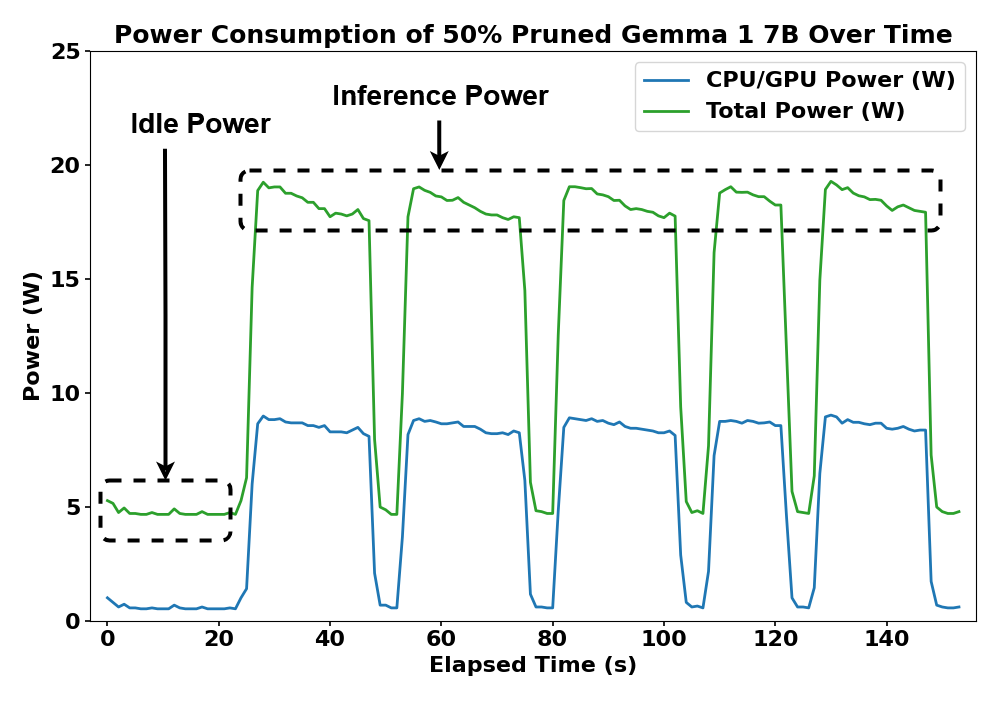}
        \vspace{-2ex}
        \footnotesize (a) Gemma 1 7B (pruned)
    \end{minipage}%
    \hfill
    \begin{minipage}[t]{0.48\linewidth}
        \centering
        \includegraphics[width=\linewidth]{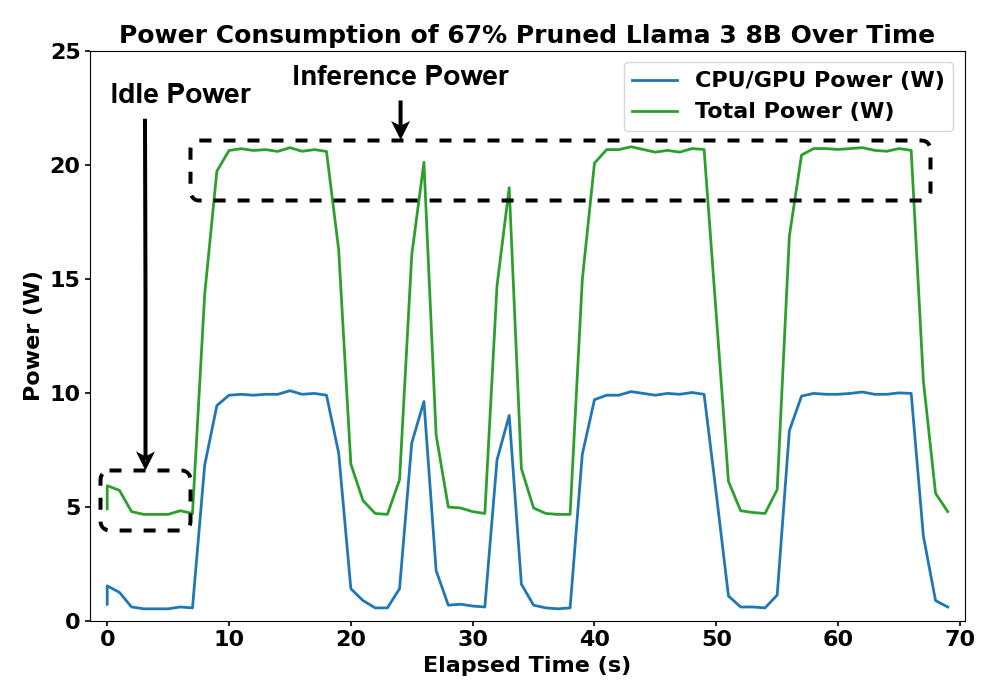}
        \vspace{-2ex}
        \footnotesize (b) Llama 3 8B (pruned)
    \end{minipage}
    \vspace{1ex}
    \caption{\footnotesize Power consumption over time on the Jetson Orin Nano. Peaks correspond to inference activity. The Gemma model has a peak power of 17.4 W while the LLaMa3 has a peak power of 18.7 W, respectively.}
    \vspace{-2ex}
    \label{fig:power_comparison}
\end{figure}

Our deployment platforms included the Raspberry Pi 5 (16\,GB RAM, CPU-only) and the NVIDIA Jetson Orin Nano Super (8\,GB RAM, Ampere GPU with unified memory). While the Raspberry Pi lacks GPU acceleration, its increased RAM allows CPU-only execution of smaller models. Figures~\ref{fig:power_comparison}(a) and (b) show power usage during five inference runs, and Table~\ref{tab:hardware_deployment_results} summarizes average power, latency, and throughput. Inference lengths varied due to differing response sizes, but the results clearly show energy spikes during inference and a performance gap tied to model size.

\begin{table}[t]
\centering
\scriptsize
\setlength{\tabcolsep}{2.5pt}
\renewcommand{\arraystretch}{1.1}
\begin{tabular}{|l|l|l|l|c|c|c|c|}
\hline
\textbf{Model} & \textbf{Prune} & \textbf{Quant} & \textbf{Device} & \textbf{Idle} & \textbf{Inf.} & \textbf{TP} & \textbf{TP/W} \\
               & (\%)           &                &                 & (W)           & Power (W)     & (tok/s)     &              \\ \hline
\multirow{2}{*}{Llama 3 8B} & \multirow{2}{*}{67} & \multirow{2}{*}{INT4} & Jetson & 5.0 & 18.7 & 22.3 & 1.2 \\ \cline{4-8}
                           &                     &                      & Pi     & 1.5 & 6.2  & 5.3  & 0.9 \\ \hline
\multirow{2}{*}{Gemma 1 7B} & \multirow{2}{*}{50} & \multirow{2}{*}{INT4} & Jetson & 5.0 & 17.4 & 11.6 & 0.7 \\ \cline{4-8}
                           &                     &                      & Pi     & 1.5 & 6.3  & 5.3  & 0.8 \\ \hline
\end{tabular}
\vspace{1mm}
\caption{\scriptsize Energy and throughput benchmarking of pruned-and-quantized LLMs on edge devices. TP = Throughput. TP/W = Throughput per Watt.}
\vspace{-10mm}
\label{tab:hardware_deployment_results}
\end{table}

\section{Conclusion}
This work proposes an on-device medical AI assistant enabled by a domain-adaptive compression framework that significantly reduces LLM size while maintaining strong performance. By leveraging input-driven saliency and post-training quantization, we show that task-specific pruning enables efficient compression with minimal accuracy loss. We validate the feasibility of lightweight, domain-specific assistants by deploying pruned and quantized models on Jetson Orin Nano (power peak of 18.7W) and Raspberry Pi 5 (power peak of 6.3W), demonstrating real-time, energy-efficient inference. In future work, we aim to extend deployment to real-world clinical settings for privacy-preserving, point-of-care applications.

\bibliographystyle{plainnat}
\bibliography{ref}

\end{document}